\documentclass[10pt,twocolumn,letterpaper]{article}

\usepackage[preprint]{cvpr}
\usepackage[most]{tcolorbox}

\usepackage{cuted}

\usepackage{currfile}

\usepackage{caption}

\usepackage[normalem]{ulem}
\makeatletter
\newcommand{\gemini@sout}[1]{{\color{red}\begingroup\let\gemini@cite\cite\DeclareRobustCommand{\gemini@wcite}[2][]{\mbox{\ifstrempty{##1}{\gemini@cite{##2}}{\gemini@cite[##1]{##2}}}}\let\cite\gemini@wcite\let\gemini@citep\citep\DeclareRobustCommand{\gemini@wcitep}[2][]{\mbox{\ifstrempty{##1}{\gemini@citep{##2}}{\gemini@citep[##1]{##2}}}}\let\citep\gemini@wcitep\let\gemini@citet\citet\DeclareRobustCommand{\gemini@wcitet}[2][]{\mbox{\ifstrempty{##1}{\gemini@citet{##2}}{\gemini@citet[##1]{##2}}}}\let\citet\gemini@wcitet\let\gemini@ref\ref\DeclareRobustCommand{\gemini@wref}[1]{\mbox{\gemini@ref{##1}}}\let\ref\gemini@wref\let\gemini@eqref\eqref\DeclareRobustCommand{\gemini@weqref}[1]{\mbox{\gemini@eqref{##1}}}\let\eqref\gemini@weqref\ifdefined\cref\let\gemini@cref\cref\DeclareRobustCommand{\gemini@wcref}[1]{\mbox{\gemini@cref{##1}}}\let\cref\gemini@wcref\fi\ifdefined\Cref\let\gemini@Cref\Cref\DeclareRobustCommand{\gemini@wCref}[1]{\mbox{\gemini@Cref{##1}}}\let\Cref\gemini@wCref\fi\protected@edef\gemini@tmp{#1}\expandafter\sout\expandafter{\gemini@tmp}\endgroup}}
\newcommand{\gemini}[2]{\ifstrempty{#1}{\ifstrempty{#2}{}{{\color{ForestGreen}#2}}}{\ifstrempty{#2}{\gemini@sout{#1}}{\gemini@sout{#1} {\color{ForestGreen}#2}}}}
\makeatother

\usepackage{listings}
\definecolor{cvprblue}{rgb}{0.21,0.49,0.74}
\usepackage[pagebackref,breaklinks,colorlinks,allcolors=cvprblue]{hyperref}
\usepackage{textcomp}
\usepackage{placeins}

\def\confName{CVPR}
\def\confYear{2027}

\title{VideoJSON: A Minimal Harness Beats Purpose-Built Memory for Ultra-Long Video Understanding}

\author{
{\large Wenqi Marshall Guo$^{1,3,\dag}$ \quad
Kuan-Hao Huang$^{2,3,\dag}$ \quad
Junda Chen$^{3}$ \quad
Hao Zhang$^{3}$} \\[4pt]
{\small $^{1}$University of British Columbia} \quad
{\small $^{2}$National Taichung University of Education \quad
$^{3}$University of California, San Diego} \\[2pt]
{\small\ttfamily wg25r@student.ubc.ca \quad
richhuang@apache.org \quad
\{junda, haozhang\}@ucsd.edu} \\
{\small$^{\dag}$ Work done during internships at UCSD.}
}

\begin{document}
\maketitle
\begin{abstract}
Answering questions over days or weeks of video requires searching for scattered clues across long timelines. Existing video agents rely on complex memory graphs and fixed retrieval pipelines, which restrict how the agent can search. We propose VideoJSON, a minimal harness for ultra-long video understanding that keeps the memory simple and lets a coding agent write its own search programs. During indexing, a vision-language model turns each minute of video into a single JSON record containing a dense caption and structured fields such as events, entities, places, and speech transcripts. At query time, the agent writes Python or Bash programs in a sandboxed container to filter, link, and read these records over multiple steps, without task-specific training or raw video access. On four ultra-long video benchmarks, VideoJSON with DeepSeek V4.1 Flash outperforms MAGIC-Video by 7.0, 20.0, and 24.75 points on EgoLifeQA, Ego-R1-Bench, and MM-Lifelong Week, and beats the best prior EgoMemReason score by 12.6 points. Across four policy models of increasing capability, VideoJSON's average score rises by up to 11.58 points over the weakest model, compared with at most 5.50 points for MAGIC-Video. Ablations show that both agentic code search and the structured JSON records drive these gains, while adding hourly summaries, extra tools, or video access changes the average score by less than 1.5 points. In addition, fine-tuning Qwen-3.5-Flash on 64 VideoJSON trajectories from an unrelated 100-hour livestream improves its MM-Lifelong scores by up to 8.25 points. Code will be open-sourced after publication. 
\end{abstract}
\section{Introduction}
\label{sec:intro}

Ultra-long video understanding (ULVU) requires answering questions whose evidence may be spread across days or weeks, as in EgoLife and MM-Lifelong~\cite{yang_egolife_2026,chen_multimodal_2026}. Consider the question, ``What did I usually do after dinner this week?'' The times of the meals determine which later activities to inspect, but the descriptions of those activities need not mention dinner. Answering the question requires locating the meals, reading descriptions from the following intervals, and comparing the activities across days.

Prior video agents organize evidence in several ways, from separate episodic, semantic, and visual stores to multimodal graphs, temporal hierarchies, and entity tables~\cite{yeo_worldmm_2026,li_bridging_2026,gong_grove_2026,rege_agentic_2026}, often with iterative retrieval or learned tool selection~\cite{tian_egor1_2025}. Others give the agent code, such as SQL over symbolic video memories~\cite{yang2024doraemongpt,fan2025videoagent} or generated programs that compose visual tools~\cite{suris2023vipergpt,subramanian2023codevqa}. However, these methods rely on purpose-built structures, including graphs, table schemas, and tool sets, that fix in advance which searches are easy to express. Recently, coding agents have become strong enough to complete long-horizon software engineering tasks with only a minimal Bash harness~\cite{mini_swe_agent,huang_deepswe_2026}. This motivates a purely agentic approach, in which a coding agent's own programs do the search.

We propose VideoJSON, a minimal harness that keeps the memory structure simple and lets a coding agent write the search. Instead of fixing in advance how evidence is linked and retrieved, VideoJSON stores the video as timestamped records and leaves every search decision to the agent's own programs, so the system gains directly from stronger coding models. It has two stages (Figure~\ref{fig:compare}). Indexing runs once before any question is seen: a vision-language model (VLM) turns each minute of video into one JSON record that holds a dense caption with nested event and entity fields. Querying runs per question: the agent writes Python or Bash programs over these records and decides from each output what to run next. For the dinner question, it can find the meals in one program and read what follows each meal in the next. The agent needs no task-specific training and no video access at query time.

In summary, our contributions are as follows:
\begin{itemize}
    \item We propose VideoJSON, a minimal harness that pairs structured per-minute JSON records with multi-step search by a coding agent.
    \item With DeepSeek V4.1 Flash, VideoJSON outperforms our MAGIC-Video rerun~\cite{li_bridging_2026} by up to 24.75 points and the best prior EgoMemReason~\cite{wang_egomemreason_2026} result by 12.6 points, while gaining up to 11.58 points over the weakest of four policy models, compared with at most 5.50 points for MAGIC-Video.
    \item Controlled ablations show that both agentic code search and structured record fields drive these gains, while added summaries, tools, or video access change the average score by less than 1.5 points.
    \item Fine-tuning Qwen-3.5-Flash on just 64 VideoJSON trajectories from an unrelated livestream improves its MM-Lifelong scores by up to 8.25 points.
\end{itemize}

\section{Related Work}
\label{sec:related}

\subsection{Long- and Ultra-Long Video Understanding}
\label{sec:related-long-video}

Answering questions about hours or days of video requires selecting evidence within limited context budgets. MovieChat consolidates visual tokens into sparse long-term memory~\cite{song2024moviechat}, while LongVU combines query-guided feature selection with spatiotemporal compression~\cite{shen2025longvu}. Video-XL summarizes video intervals into compact key--value representations for hour-scale understanding~\cite{shu2025videoxl}.

Multiple works approach this problem by storing video-derived evidence in external memory and retrieving it on demand. DrVideo iteratively augments textual video documents with details from selected frames~\cite{ma2025drvideo}. WorldMM maintains separate episodic, semantic, and visual memories, with multiscale episodic retrieval~\cite{yeo_worldmm_2026}. MAGIC-Video combines query-seeded diffusion over a multimodal graph with narrative-chain retrieval for entities and activities~\cite{li_bridging_2026}. GROVE organizes streaming observations into perceptual records, moments, episodes, and recurring cross-day patterns, with retrieval skills matched to each level~\cite{gong_grove_2026}. Like VideoJSON, MERIT~\cite{choi_keep_2026} uses a simple, queryable index instead of heavyweight, query-agnostic memory construction, storing information under dedicated keys such as events, dialogue, objects, and a summary. However, its retrieval still relies on similarity scoring and neighborhood filtering, rather than allowing the model to query the index freely.

\begin{figure*}[h]
    \centering
    \includegraphics[width=\linewidth]{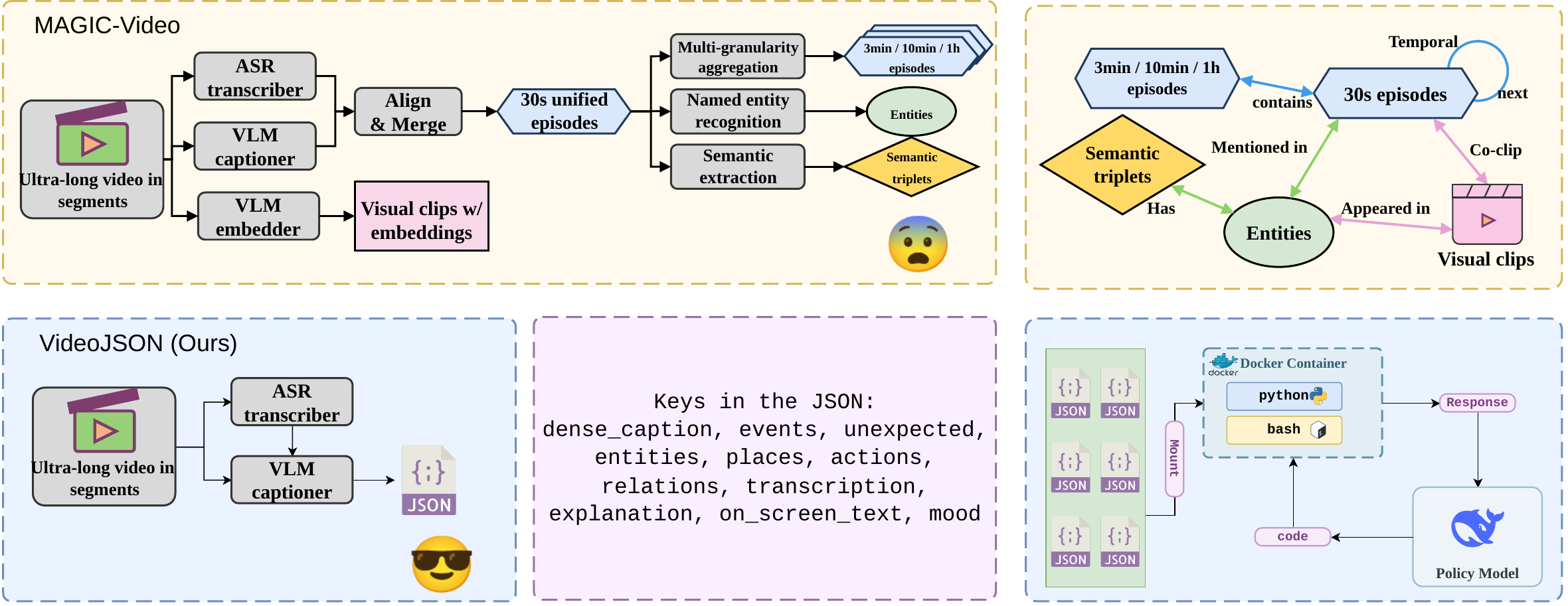}
    \caption{Indexing pipelines and index structures of MAGIC-Video (top) and VideoJSON (bottom). MAGIC-Video builds a graph of multi-granularity episodes, entities, semantic triplets, and visual clips through several stages of LLM calls and embeddings. VideoJSON captions each minute of video into one JSON record with the listed keys. At query time (bottom right), the policy model writes Python or Bash code, which runs in a Docker container with the records mounted read-only, and reads the output.}
    \label{fig:compare}
\end{figure*}

Several video agents use SQL to query stored evidence: VideoAgent and DoraemonGPT query object records and symbolic video memories, respectively~\cite{fan2025videoagent,yang2024doraemongpt}, while EGAgent queries an entity table containing time-stamped entity relations, with visual and audio search~\cite{rege_agentic_2026}. VideoJSON also queries stored evidence with agent-written code, but over nested JSON records instead of relational tables. 

\subsection{Database-Based RAG and Agent Memories}
\label{sec:related-sql-rag}
Recent work treats retrieval and agent memory as data-management problems. SAG stores document chunks as events and entities in relational tables, and shared entities act as SQL join keys at query time, so related events are reached through joins without materializing a global knowledge graph~\cite{wu_sag_2026}. Orogat and Mansour frame persistent agent memory as a data-management workload, in which ingestion, revision, forgetting, and retrieval all operate on a state that evolves over time~\cite{orogat_agent_2026}. A-MEM stores each memory as a structured note with context, keywords, and tags, links new notes to related earlier ones, and updates existing notes as new memories arrive~\cite{xu_amem_2025}.

VideoJSON treats its per-minute JSON record collection as a memory store and uses the agent's Python or Bash programs as its query interface. The store is fixed before any question arrives; the programs are written per question.

\subsection{Minimal Harnesses}
Recent coding agents keep the harness around the model small and let the model's own programs do the work. CodeAct makes executable Python the action, with the interpreter's output fed back to the model for its next step~\cite{wang_codeact_2024}. mini-SWE-agent gives the model a Bash-only action space and an append-only message history~\cite{mini_swe_agent}. Recursive Language Models treat an input too long for the context window as part of the external environment, which the model inspects and decomposes with code before calling a language model over selected pieces~\cite{zhang_rlm_2025}. VideoJSON follows this direction for video memory: the record collection stays in the execution environment, while the agent's programs select what to read from it.

\section{VideoJSON}
\subsection{Indexing}
Figure~\ref{fig:compare} gives an overview of VideoJSON, which stores a video collection as one JSON record per minute. Each record carries a dense caption of the clip together with structured fields that a program can filter on: events with start and end seconds within the minute, entities, places, actions, relations, unexpected moments, transcription, on-screen text, mood, and an explanation field. The caption fields are the same for every source we evaluate, including egocentric recordings, gameplay, and livestreams; the record metadata identifies the clip or recording, the day when available, and the start and end time of the minute.

We use a VLM (Qwen-3.8-Flash-Next~\cite{qwen38}) to caption each 60-second clip independently so clips can be captioned in batches. For each clip, the VLM receives its frames at 1 fps and a speech transcript from Qwen-3-ASR-1.7B~\cite{shi_qwen3asr_2026}; for the EgoLife recordings, it also receives the transcripts released with the dataset. The memory is built once per video collection before any question is seen, and each question reuses this prebuilt memory, with the records a question may read restricted as described in Section~\ref{sec:settings}. We do not use a cascaded hierarchy (e.g., hourly or daily indices above clip-level captions), nor do we adopt the multi-table indexing structure of MAGIC-Video (e.g., entity tables, event chains, semantic triplets). We evaluate both in the ablations.

Our indexing pipeline and index structure are considerably simpler than those of MAGIC-Video, which relies on sequential LLM calls and embeddings to build a complicated graph database. This sequential pipeline is difficult to parallelize. Figure~\ref{fig:compare} contrasts both pipelines and index structures.

\subsection{Agentic JSON Queries}
The agent answers a question by writing programs over these records, using two execution tools. \texttt{exec} runs a Python or Bash program in a clean Docker container assigned to the question and returns its output. \texttt{exec\_qa} runs the program the same way but, instead of returning the output, passes the code, the full output, and a question from the main agent to a separate question-answering call, and returns only the answer. 

Calls can be iterative. The agent reads an output, decides what to inspect next, and writes the next program, so a retrieval procedure can be built up over several steps when one program is not enough. A program can select records by field values or by time, compare timestamps across records, group and count, or print selected captions in full. In the minimal configuration, the agent reads only the caption memory and has no access to the raw video or the network; Table~\ref{tab:ablation} adds a \texttt{video\_qa} tool as a separate setting. When using \texttt{video\_qa}, the agent can select a window of video and query the policy model on that window, without the video entering the agent's context. Each question has a budget of 30 model calls, counting the agent's calls and any \texttt{video\_qa} calls, after which the agent must answer. The system prompt is detailed in Appendix~\ref{sec:prompt}.

Take the question from the introduction, ``What did I usually do after dinner this week?'' A first program can scan the week's records and print the day and end time of every event whose description mentions dinner. From those timestamps, a second program can print the events and dense captions from the records that follow each dinner, with the agent choosing how far forward to read. A third program can then check a candidate activity, for example by counting the days on which it appears after dinner.

\section{Experimental Setup}

\begin{table*}[t]
\centering
\small
\begin{tabular}{llccccc}
\toprule
Method &  Model&EgoLifeQA& Week & Day & Month  &Ego-R1-Bench\\
\midrule
ReMA \cite{chen_multimodal_2026}&  GPT-5&-- & 18.82$^{*}$ & 16.75$^{*}$ & 18.62$^{*}$  &--\\
GROVE \cite{gong_grove_2026} &  GPT-5.2&-- & 22.75$^{*}$ & 23.50$^{*}$ & 19.98$^{*}$  &--\\
\midrule
Ego-R1 \cite{tian_egor1_2025} &  GPT-OSS-120B + Qwen-3.5-Flash&37.6$^{\dagger}$ & 14.25$^{\dagger}$ & 6.50$^{\dagger}$ & --  & 30.0$^{\dagger}$\\
AVP \cite{wang_active_2026} &  GPT-OSS-120B + Qwen-3.5-Flash&30.2$^{\dagger}$ & 14.50$^{\dagger}$ & 4.25$^{\dagger}$ & --  &28.7$^{\dagger}$\\
WorldMM \cite{yeo_worldmm_2026} &  Qwen-3.5-Flash&56.0$^{\dagger}$ & 31.42$^{\dagger}$ & 8.50$^{\dagger}$ & --  &57.3$^{\dagger}$\\
MAGIC-Video \cite{li_bridging_2026} &  Qwen-3.5-Flash&67.6*& 30.92$^{\dagger}$ & 10.08$^{\dagger}$ & 24.5$^{*}$  &64.7*\\
MAGIC-Video \cite{li_bridging_2026} &  Qwen-3.5-Flash&68.6& 26.25& --& --&64.0\\
GEB \cite{ren_timeline_2026} &  Qwen-3.5-Flash&\textbf{72.0}$^{\dagger}$ & 36.83$^{\dagger}$ & 17.58$^{\dagger}$ & --  &71.3$^{\dagger}$\\
VideoJSON (Ours) &  Qwen-3.5-Flash&65.8 & \textbf{41.50}& \textbf{34.75} & \textbf{33.15}  &\textbf{74.0}\\
\midrule
MAGIC-Video \cite{li_bridging_2026}&  DeepSeek &	70.4& 35.00& --& --&58.0\\
VideoJSON (Ours) &  DeepSeek &\textbf{77.4}& \textbf{59.75} & \textbf{64.75} & \textbf{48.39} &\textbf{78.0}\\
\bottomrule
\end{tabular}
\caption{Main results with Qwen-3.5-Flash and DeepSeek. $^{*}$ denotes scores copied from the original paper; $^{\dagger}$ denotes scores copied from GEB, a concurrent preprint. Unmarked MAGIC-Video scores are our reruns with its released index on EgoLifeQA, MM-Lifelong Week, and Ego-R1-Bench. EgoLifeQA and Ego-R1-Bench use only the A1\_JAKE subset.}
\label{tab:main}
\end{table*}

\subsection{Benchmarks and Protocol}\label{sec:settings}
We evaluate on four ULVU benchmarks: EgoLifeQA~\cite{yang_egolife_2026}, Ego-R1-Bench~\cite{tian_egor1_2025}, MM-Lifelong~\cite{chen_multimodal_2026}, and the newer EgoMemReason~\cite{wang_egomemreason_2026}.

EgoLifeQA, Ego-R1-Bench, and EgoMemReason use the same set of egocentric recordings from EgoLife. Following MAGIC-Video, we use the A1\_JAKE subset for EgoLifeQA (500 questions) and Ego-R1-Bench (50 questions), while using the full test set for EgoMemReason. For each question, the caption memory and any video access are restricted to the corresponding wearer's history up to the query time. Because Ego-R1-Bench writes people's names in Chinese differently from EgoLifeQA, we also provide its questions and options in both English and Chinese. 

MM-Lifelong provides open-ended questions. We use the Week and Day test splits and the Month validation split, because the Month subset has no test split. The caption memory covers the full recording collection for each setting. In contrast, the released evaluator from MAGIC-Video loads only the videos named in each Month question's ground-truth clue annotations, which makes the task easier. We report its published Month score for context only. MAGIC-Video's released index does not cover the full Month collection, and its paper does not report Week or Day. For Week and Day, Table~\ref{tab:main} uses the scores reported in Grounded Entity Biographies (GEB)~\cite{ren_timeline_2026}. We rerun MAGIC-Video with its released index on EgoLifeQA, Ego-R1-Bench, and MM-Lifelong Week (which uses the same videos as EgoLifeQA). With Qwen-3.5-35B-A3B~\cite{qwen3.5} (hereafter Qwen-3.5-Flash), our reruns score 68.6 on EgoLifeQA and 64.0 on Ego-R1-Bench, close to the reported 67.6 and 64.7.

\begin{figure}
    \centering
    \includegraphics[width=0.45\linewidth]{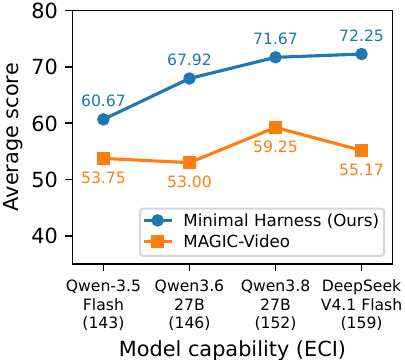}
    \includegraphics[width=0.45\linewidth]{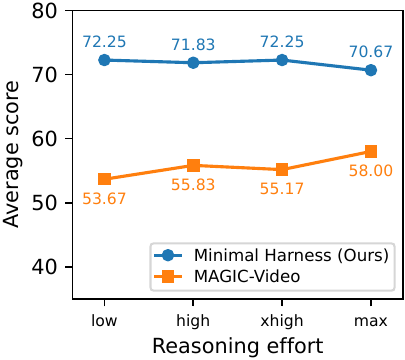}
    \caption{VideoJSON vs.\ MAGIC-Video scaling with the model's capability and reasoning level. Scores are averaged over the 200-question EgoLifeQA subset, Ego-R1-Bench, and MM-Lifelong Week.}
    \label{fig:scaling}
\end{figure}

For the ablation, we use the full Week and Day test sets (200 questions each), alongside fixed 200-question subsets of EgoLifeQA A1\_JAKE and the Month validation split. For the scaling experiments (Figure~\ref{fig:scaling}), we use the same EgoLifeQA subset, the Week test set, and Ego-R1-Bench.

EgoLifeQA, Ego-R1-Bench, and EgoMemReason are scored by option accuracy. For MM-Lifelong, GPT-5 \cite{singh2025openai} assigns each answer a score from 0 to 5; we map scores of 4 and 5 to 1, a score of 3 to 0.5, and lower scores to 0, and report the mean multiplied by 100. The judge model and prompt are consistent with the original MM-Lifelong paper \cite{chen_multimodal_2026} and other papers' benchmark settings. On EgoLifeQA, the released English translations contain mistranslations and inconsistencies. For example, the English text translates Baked Wheaten Cake (\textit{huoshao}, a northern Chinese snack) as ``Fire Burn'', and for ``I put the pot aside, what was in the pot before?'', option C is rice in Chinese but puzzle in English. VideoJSON therefore uses only the Chinese question and option text. Since the index of MAGIC-Video is in English, providing only Chinese would harm performance; we therefore provide MAGIC-Video with both English and Chinese, and report VideoJSON with bilingual questions and options as well for comparison in Appendix~\ref{sec:bilingual}. 

\subsection{Baselines and Models}
For baselines, we use the results reported in GEB and in the original papers, and run our method with the policy model used in GEB, Qwen-3.5-Flash.

We hypothesize that stronger models benefit more from a minimal harness, and therefore also run our method and MAGIC-Video with DeepSeek V4.1 Flash 1017~\cite{deepseek-ai_deepseekv41flash_2026} (hereafter DeepSeek) at xhigh reasoning effort. To test how each harness scales with model capability, we run both methods with two additional policy models (Qwen-3.6-27B and Qwen-3.8-27B) and with DeepSeek at four reasoning levels (low, high, xhigh, max). VideoJSON uses Qwen-3.8-Flash-Next for captioning. MAGIC-Video uses its released index, except in runs that also need Day and Month (Table~\ref{tab:qwen35_captioning}), which use an index rebuilt with Qwen-3.8-Flash-Next. Implementation details and inference engine configurations are included in the source code.

\section{Results}
\subsection{Main Results}

Table~\ref{tab:main} shows the main results with Qwen-3.5-Flash and DeepSeek. With Qwen-3.5-Flash, VideoJSON leads prior methods on every dataset except EgoLifeQA, outperforms the concurrent GEB on 3 of the 4 shared datasets (nearly doubling its Day score), and beats systems built on much stronger proprietary models such as ReMA (GPT-5) and GROVE (GPT-5.2). Upgrading the policy model to DeepSeek improves VideoJSON on all five datasets, outperforming MAGIC-Video on every shared dataset. To measure the variation and stability of our method, we run 5 trials and report the standard deviation in Appendix~\ref{sec:multi_runs}. In contrast, MAGIC-Video with DeepSeek improves on EgoLifeQA and Week but drops on Ego-R1-Bench, and on both Week and Ego-R1-Bench it scores below VideoJSON with Qwen-3.5-Flash.
\subsection{Scaling with Model Capability}
We evaluate VideoJSON and MAGIC-Video across four policy models: Qwen-3.5-Flash \cite{qwen3.5}, Qwen-3.6-27B \cite{qwen3.6-27b}, Qwen-3.8-27B \cite{qwen38_27b}, and DeepSeek V4.1 Flash 1017, in increasing order of Epoch Capabilities Index (ECI) \cite{ho2025rosetta}. We use the average score on EgoLifeQA, Ego-R1-Bench, and MM-Lifelong Week, the datasets on which MAGIC-Video can run with its released index. Figure~\ref{fig:scaling} (a) shows the scaling results, where the x-axis is the model's ECI (labeled with the model name) and the y-axis is the average score over the three datasets. Both methods generally improve as the model gets stronger. VideoJSON outperforms MAGIC-Video with every model. Even with the weakest model, Qwen-3.5-Flash, VideoJSON outperforms MAGIC-Video with the strongest model. VideoJSON also gains more from stronger models: compared with Qwen-3.5-Flash, VideoJSON gains 11.58 points at its best point (DeepSeek V4.1 Flash), while MAGIC-Video gains only 5.50 points at its best point (Qwen-3.8-27B).

Besides model variants, we also test scaling with reasoning effort (Figure~\ref{fig:scaling} (b)). From low to max effort, VideoJSON stays within 1.6 points, while MAGIC-Video gains 4.3 points but remains below VideoJSON's low-effort score.

\subsection{EgoMemReason Results}
Table~\ref{tab:egomemreason} compares VideoJSON, using DeepSeek, with the results reported in the EgoMemReason paper~\cite{wang_egomemreason_2026}. VideoJSON leads or ties in 3 of the 6 categories and is 12.6 points above the best prior overall score. Its largest margin is on Ordering (86.0 vs.\ 44.0), and it underperforms most on Counting (32.0 vs.\ 50.0).

\begin{table*}[!t]
\centering
\small
\setlength{\tabcolsep}{3.8pt}
\begin{tabular}{lllccccccc}
\toprule
Method   &Backbone&Modality& Tracking & Counting & Ordering & Linking & Spatial & Activity & Overall \\
\midrule
Direct Model   &Gemini-3-Flash   &frames-only& \underline{46.0} & 28.0 & 36.0 & \textbf{44.0} & \underline{44.0} & \underline{44.0} & \underline{39.6} \\
Direct Model   &Gemini-3.1-Pro   &frames-only& 40.0 & 26.0 & \underline{44.0} & 33.0 & 40.0 & \textbf{48.0} & 37.4 \\
AVP   &Gemini-3-Flash   &frames-only& 34.0 & \underline{42.0} & 31.0 & 27.0 & 38.0 & 34.0 & 34.0 \\
Best Prior& -    &-& \underline{46.0}& \textbf{50.0}& \underline{44.0}& \textbf{44.0}& \textbf{50.0}& \textbf{48.0}& \underline{39.6}\\
VideoJSON (Ours)  &DeepSeek&caption-only& \textbf{53.0} & 32.0 & \textbf{86.0} & \underline{43.0} & \textbf{50.0} & \underline{44.0} & \textbf{52.2} \\
\bottomrule
\end{tabular}
\caption{Accuracy (\%) on the 500-question EgoMemReason test set. Reference results come from the original benchmark paper~\cite{wang_egomemreason_2026}. The Best Prior row takes the highest score in each category from the full table of the original paper. The best result in each column is in bold and the second best is underlined.}
\label{tab:egomemreason}
\end{table*}

\subsection{Ablation}
Because our harness is already minimal, we perform a reverse ablation by adding three components: (1) the \texttt{video\_qa} tool; (2) a decision model API (Laya~\cite{m_laya_2026}) to filter or classify captions; and (3) an hourly caption above the clip-level captions that aggregates information within each hour. We also test our harness with MAGIC-Video's index, rebuilt with Qwen-3.8-Flash-Next and converted into JSON (w/ MAGIC Index), and with our JSON records replaced by dense captions (Dense Caption Only). Finally, we test a setting in which the agent is given a single simple BM25 search tool to search our VideoJSON index. 

Table~\ref{tab:ablation} shows the ablation results. Adding \texttt{video\_qa} gives the largest gains on Week, Day, and Month but lowers EgoLifeQA by 2 points, and more than doubles the cost per question. Exposing the decision API improves Day and Month but lowers EgoLifeQA and Week, while adding a separate model service to maintain. Adding hourly captions barely changes the results, so the clip-level captions already give the agent what it needs.

Replacing VideoJSON records with dense captions or with MAGIC-Video's memory lowers the average score without reducing cost. Dense captions alone lower EgoLifeQA, Week, and Month but slightly raise Day. On MM-Lifelong Week, querying MAGIC-Video's memory with our coding agent already gains a large improvement over MAGIC-Video (Table~\ref{tab:main}).

Using a single BM25 search tool substantially increases the cost and lowers the scores. This is likely due to a large amount of returned data. 

\begin{table}[!t]
\centering
\small
\setlength{\tabcolsep}{6pt}
\resizebox{\linewidth}{!}{
\begin{tabular}{lccccc}
\toprule
Setting & EgoLifeQA & Week & Day & Month & Cost/q (US\textcent)  \\
\midrule
Minimal Harness & \textbf{79.00} & \underline{59.75} & 64.75 & 49.50 & 1.52  \\
+ \texttt{video\_qa} & 77.00 & \textbf{60.50} & \textbf{67.25} & \textbf{53.75} & 3.57  \\
+ Decision API & 77.00 & 58.50 & 66.25 & \underline{52.25} & 1.58  \\
+ Hourly Caption & \underline{78.50} & \underline{59.75} & 64.50 & 51.75 & \underline{1.50}  \\
w/ MAGIC Index & 71.50 & 50.25 & 27.00 & 47.50 & 1.64  \\
Dense Caption Only& 77.50 & 52.00 & \underline{66.50} & 45.00 & 1.58  \\
BM25 Only& 64.50& 52.50& 	49.75& 	43.25& 2.61\\
\bottomrule
\end{tabular}}
\caption{Ablation on 200 questions per dataset with DeepSeek V4.1 Flash 1017 (xhigh reasoning). Each $+$ row adds one component without stacking additions. The w/ MAGIC Index rows use the MAGIC-Video index rebuilt with Qwen-3.8-Flash-Next instead of the released index used in Table~\ref{tab:main}. Costs are reported at \$0.15/M uncached input, \$0.003/M cached input, and \$0.60/M output tokens based on DeepSeek's official on-peak price. The best result in each column is in bold and the second best is underlined.}
\label{tab:ablation}
\end{table}

\subsection{Indexing Wall Time}
\label{sec:wall}
\begin{figure*}
    \centering
    \includegraphics[width=\linewidth]{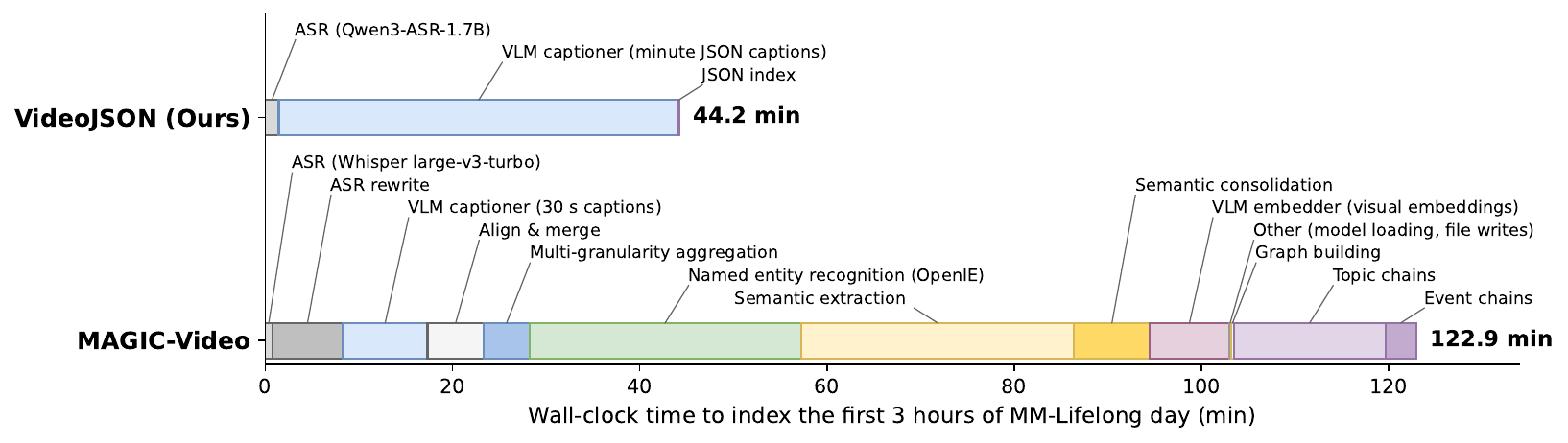}
    \caption{Index time for VideoJSON and MAGIC-Video}
    \label{fig:wall}
\end{figure*}

Measuring indexing wall-clock time is challenging because it depends heavily on workflow optimization. This is particularly true for MAGIC-Video, where inter-index dependencies limit parallelization. Nonetheless, we can provide an approximate comparison based on overall runtime. To ensure a fair evaluation, we ran an optimized version of the official MAGIC-Video pipeline configured for higher concurrency within stages and sequential across stages, while leaving our inherently parallelizable method unchanged. We used a concurrency of 16 in both experiments. Using Qwen-3.5-Flash on a single B200 GPU with vLLM \cite{kwon_efficient_2023}, we captioned 3 hours of video (3 complete processing units of MAGIC-Video) from the MM-Lifelong Day benchmark. As shown in Figure~\ref{fig:wall}, our method is significantly faster than MAGIC-Video.

\subsection{Code and Trace Analysis}
To understand how the agent answers each question, we analyze what its code does (code nature), which language it uses, which keys it reads, how often it calls \texttt{exec\_qa}, and how many tokens it uses. For code nature and keys, we use DeepSeek to classify each piece of code as regex or search, reading a specific interval in detail, or other, and to extract the keys it reads. For language, we count the language field, and Python run in a Bash environment is counted as Bash. Token consumption and the \texttt{exec\_qa} share are calculated directly from the traces.

\begin{figure}[t]
\centering
\includegraphics[width=\linewidth]{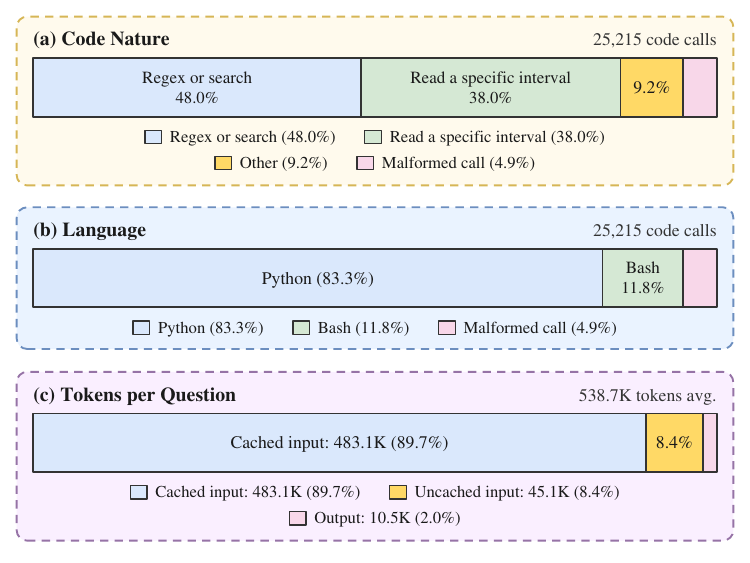}
\caption{(a) Code nature of each code piece. (b) Language of each code piece. (c) Tokens per question of the DeepSeek agent, averaged over the five datasets. Cached input assumes a perfect prefix cache: each agent call's cached input is the previous call's input and output.}
\label{fig:token-pie}
\end{figure}

\paragraph{Code nature.}
Figure~\ref{fig:token-pie} (a) shows what the agent's code does, averaged over the five datasets. Most code either searches the index (48.0\%), with substring tests, keywords, or regular expressions over the minute records, or reads a specific interval in detail (38.0\%), picking records by clip ID, list position, or time range and printing their caption fields. Other code inspects the files or the structure of the data, counts records, or is empty (9.2\%); 4.9\% of the calls are malformed, with the code placed in the \texttt{language} argument. By language (Figure~\ref{fig:token-pie} (b)), 83.3\% of calls use Python and 11.8\% use Bash.

\paragraph{Keys.}
Table~\ref{tab:keys} lists the ten most used keys, with shares averaged over the five datasets. Most code reads \texttt{clip\_id} (89.2\%), which the agent uses to locate records and to cite them in later calls. Among the caption fields, \texttt{dense\_caption} (60.9\%) and \texttt{transcription} (50.1\%) dominate, followed by \texttt{events}, \texttt{entities}, and \texttt{explanation}. On EgoLife, the wall-clock keys \texttt{start\_time} and \texttt{day} are read by about 90\% and 70\% of the code, since questions are tied to a query time. In 18.7\% of the code, the agent stringifies a whole record or caption and searches the string instead of choosing fields.

\begin{table}[t]
\centering
\small
\resizebox{\linewidth}{!}{
\begin{tabular}{lcccccc}
\toprule
Key & EgoLifeQA & Week & Day & Month & Ego-R1-Bench & Avg. \\
\midrule
\texttt{clip\_id} & 83.8 & 89.8 & 87.4 & 93.1 & 91.6 & 89.2 \\
\texttt{dense\_caption} & 59.6 & 60.3 & 56.7 & 66.3 & 61.7 & 60.9 \\
\texttt{start\_time} & 91.2 & 88.3 & -- & -- & 92.1 & 54.3 \\
\texttt{transcription} & 56.1 & 47.5 & 30.8 & 51.3 & 64.8 & 50.1 \\
\texttt{day} & 72.1 & 66.0 & -- & -- & 74.1 & 42.5 \\
\texttt{events} & 36.4 & 32.6 & 30.3 & 34.0 & 32.0 & 33.1 \\
\texttt{end\_time} & 47.5 & 34.2 & -- & -- & 49.0 & 26.2 \\
\texttt{entities} & 27.6 & 19.9 & 17.3 & 15.4 & 24.1 & 20.9 \\
\texttt{explanation} & 12.2 & 20.0 & 24.9 & 21.5 & 19.0 & 19.5 \\
whole JSON & 13.4 & 12.9 & 30.5 & 20.4 & 16.4 & 18.7 \\
\bottomrule
\end{tabular}
}
\caption{Share of code (\%) that reads or searches each key, top 10 by the average over the five datasets. \texttt{start\_time}, \texttt{day}, and \texttt{end\_time} exist only in the EgoLife index (--), and Avg.\ counts -- as 0. Whole JSON: the code stringifies a whole record or caption.}
\label{tab:keys}
\end{table}

\paragraph{\texttt{exec} and \texttt{exec\_qa}.}
The agent almost always reads the raw output itself (Table~\ref{tab:exec-qa}): \texttt{exec\_qa} accounts for 0.72\% of the code calls on average. This low usage suggests that the harness can be simplified further (\ifcsname r@sec:no_exec_qa\endcsname Section~\ref{sec:no_exec_qa}\else Section~A\fi\ of the supplementary material). 

\begin{table}[t]
\centering
\small
\begin{tabular}{lc}
\toprule
Tool & Share (\%)\\
\midrule
\texttt{exec} & 99.28 \\
\texttt{exec\_qa} & 0.72 \\
\bottomrule
\end{tabular}
\caption{Share (\%) of \texttt{exec} and \texttt{exec\_qa} among the code calls.}
\label{tab:exec-qa}
\end{table}

\paragraph{Token consumption.}
Figure~\ref{fig:token-pie} (c) breaks down the tokens per question, averaged over the five datasets. Cached input makes up 89.7\% (483.1K tokens), uncached input 8.4\% (45.1K), and output 2.0\% (10.5K). Because each call re-reads the growing conversation, the prefix is reused across calls, while the new input and output stay small.

\subsection{Captioning Model Capability}
We choose Qwen-3.8-Flash-Next as the captioning model since it is fast, inexpensive, and compact enough to deploy on our servers. However, some users may need to run captioning locally because of connectivity and privacy constraints. We therefore evaluate our approach using captions produced by Qwen-3.5-Flash, the VLM captioner MAGIC-Video uses on MM-Lifelong. To ensure a consistent comparison, we use Qwen-3.8-27B as the policy model, as it can also run locally while offering stronger agentic capabilities. As Table~\ref{tab:qwen35_captioning} shows, even with a weaker captioner, performance remains strong and meaningfully higher than MAGIC-Video, whose released index is built with GPT-OSS-120B from the captions released with EgoLife. Thus, the gains come mainly from our approach rather than from the captioning model. When we rebuild MAGIC-Video's index with Qwen-3.8-Flash-Next, its performance drops by 11.0 points on EgoLifeQA and 3.8 on Week, likely because the additional data in the index makes the graph more complex.

\begin{table*}[t]
\centering
\small
\begin{tabular}{llccccc}
\toprule
Method & Captioning Model & EgoLifeQA & Week & Day & Month & Ego-R1-Bench \\
\midrule
MAGIC-Video &  EgoLife captions + GPT-OSS-120B &72.50 & 41.25 & -- & -- & 64.00 \\
MAGIC-Video &  Qwen-3.8-Flash-Next & 61.50 & 37.50 & 10.50 & 30.50 & -- \\
VideoJSON (Ours)&  Qwen-3.8-Flash-Next&77.00 & 58.00 & 61.00 & 49.25 & 80.00 \\
VideoJSON (Ours)&  Qwen-3.5-Flash&74.50 & 57.75 & 49.00 & 48.00 & 74.00 \\
\bottomrule
\end{tabular}
\caption{Results with Qwen-3.8-27B as the policy model and different captioning models, on 200 questions per dataset (50 for Ego-R1-Bench). The first MAGIC-Video row uses its released index.}
\label{tab:qwen35_captioning}
\end{table*}

\begin{table}[t]
\centering
\small
\setlength{\tabcolsep}{3.5pt}
\begin{tabular}{lccc}
\toprule
Model & Week & Day & Month \\
\midrule
Qwen-3.5-Flash & 38.50 & 31.50 & 34.35 \\
+ SFT & \textbf{46.75} & \textbf{34.75} & \textbf{39.73} \\
\midrule
$\Delta$ & +8.25 & +3.25 & +5.38 \\
95\% CI & [1.25, 15.25] & [$-$3.00, 9.50] & [1.85, 8.91] \\
\bottomrule
\end{tabular}
\caption{Qwen-3.5-Flash before and after fine-tuning on 64 MediaStorm-Livestream trajectories, on identical MM-Lifelong questions scored by GPT-5. $\Delta$ and its 95\% bootstrap interval are paired over questions.}
\label{tab:sft}
\end{table}

\section{Fine-tuning on VideoJSON Trajectories}\label{sec:sft}
While VideoJSON requires no task-specific training, every query produces a multi-step search trajectory that can be distilled into a smaller model, which is useful for applications such as personal assistants on local or edge devices. This section tests whether VideoJSON trajectories collected on unrelated footage can improve such a model.

Training data for ultra-long video understanding is scarce because it is hard to build, and existing datasets overlap with our evaluation benchmarks. Ego-R1's training data~\cite{tian_egor1_2025} covers the same EgoLife recordings as EgoLifeQA. The MM-Lifelong Month development split~\cite{chen_multimodal_2026} is separate from Week and Day but shares videos with the Month validation split, so training on it leaves the 100-hour Month setting unusable for testing. We therefore build a question set on a separate livestream, collect VideoJSON trajectories on it with a larger model, and train Qwen-3.5-Flash, a small mixture-of-experts model, on them with supervised fine-tuning (SFT).

\subsection{MediaStorm-Livestream}
MediaStorm-Livestream is built from the public Bilibili replay of a 100-hour snow-survival livestream produced by MediaStorm.\footnote{\url{https://www.bilibili.com/video/BV1rKzPBVEq1/}} The replay has 65 parts totaling 101.2 hours, with Chinese speech and on-screen text. We build JSON records over 30-second clips instead of 60-second ones, using Qwen-3.8-27B~\cite{qwen38_27b} for the captions and Qwen-3-ASR-1.7B~\cite{shi_qwen3asr_2026} for speech.

The set has 400 open-ended questions in MM-Lifelong's format, each with a clue interval and a question type. Because the set is built for training, we generate it automatically. LLM agents locate candidate events through the captions and, for a few questions, viewers' bullet comments (\textit{danmaku}). They then write each question and answer from frames decoded from the original video, including on-screen subtitles, and list the facts a correct answer must contain, together with the frames that show them.

As a blind check, Qwen-3.8-Flash-Next answers each question from the frames without seeing the answer, and its answer is compared with the reference. The questions are then reviewed against the frames by three GPT-6.1-Sol reviewers (xhigh effort) and five GPT-6-Astra reviewers (max effort); 32 of 432 drafted questions are rejected, and 7 single-fact questions are replaced with harder ones. The final 400 questions cover 61 of the 65 parts, and their types follow MM-Lifelong's categories, led by Language Content Recall (147), Event Tracking (73), and Event Recognition (52). No video or question overlaps with the evaluation benchmarks.

\subsection{Trajectory Collection and SFT Distillation}
We split the questions by time. Questions whose clue intervals are less than 10 minutes apart go into the same group. Of the resulting 112 groups, 78 (306 questions, excluding one flagged as inconsistent) are used for training, and the rest are held out. Qwen-3.8-Flash-Next answers all 400 questions with VideoJSON at low reasoning effort and receives full credit on 356 of them, as judged by Qwen-3.8-Flash-Next itself. We use Qwen-3.8-Flash-Next instead of GPT-5 as the judge in this case because it is used to curate training data, and we want to avoid leaking the final judge's preference. We enable the \texttt{video\_qa} tool here and in the evaluation below so that the distilled model learns both caption search and targeted frame inspection.

Going through the training questions in a fixed hashed order, we take the first 64 trajectories that receive full credit, contain no tool errors, and fit within 16K tokens per training example. The trajectories record the model's reasoning, programs, tool outputs, and final answer, and 45 of the 64 use \texttt{video\_qa}. Each assistant turn, with its reasoning and tool calls, becomes one training example, for 352 examples in total. The context of each example contains all earlier turns, including their reasoning, and the loss covers only the current turn.

We fine-tune Qwen-3.5-Flash with LoRA~\cite{hu_lora_2021} on its attention and shared-expert projections, keeping the router and routed experts frozen (rank 16, $\alpha=32$, dropout 0.05; 21.2M trainable parameters). Training runs for one epoch (88 steps) with AdamW \cite{loshchilov_decoupled_2019}, a constant learning rate of $5\times10^{-5}$, and an effective batch size of 4.
\subsection{Results}
On the Week and Day test sets and the Month validation split (200, 200, and 623 questions), the base and fine-tuned models answer the same questions with the same sampling parameters (temperature 1.0, top-$p$ 0.95, top-$k$ 20). In both runs, the base model answers the \texttt{video\_qa} calls. Answers are scored by GPT-5 with the judge prompt and score mapping of Section~\ref{sec:settings}. Table~\ref{tab:sft} reports paired differences with 95\% bootstrap intervals over questions. 

Fine-tuning raises Week from 38.50 to 46.75, Month from 34.35 to 39.73, and Day from 31.50 to 34.75 (Table~\ref{tab:sft}). The fine-tuned model calls \texttt{video\_qa} on 44--58\% of questions, compared with 16--30\% for the base model.

\section{Conclusion}
A coding agent can write its own search over weeks of video. With one JSON record per minute and a minimal two-tool harness, it outperforms MAGIC-Video with all four policy models, and the margin tends to grow as the model gets stronger. Both the query procedure and the record content matter: our agent already achieves reasonable performance with MAGIC-Video's memory, and VideoJSON records raise the score further. The design lesson is to put effort into what each record contains and let the model write the search, so the system improves as the model does. The harness also keeps every search as a trajectory, and Section~\ref{sec:sft} shows that such trajectories improve a small model. A natural next step is to train agents on them to tell when the captions are not enough, and the video is worth reading. 

\section*{LLM Usage Statement}
LLMs are part of our method and evaluation as described in the paper, and Codex (OpenAI) agents wrote the MediaStorm-Livestream questions. We used Claude Code (Anthropic) and Codex as coding agents for the experiments and Claude (Anthropic) and Gemini (Google) to edit the writing. We also used CSPaper and ARD-Review \cite{guo_ardreview_2026a} to obtain feedback and improve the paper. 
\FloatBarrier
{
    \small
    \bibliographystyle{ieeenat_fullname}
    \bibliography{main,related_long_video}
}

\clearpage
\setcounter{page}{1}
\maketitlesupplementary
\appendix
\section{Without the \texttt{exec\_qa} Tool}
\label{sec:no_exec_qa}
In the code and trace analysis (Table~\ref{tab:exec-qa}), we find that the agent rarely uses the \texttt{exec\_qa} tool. In this section, we test whether it can be removed completely. When an \texttt{exec} output exceeds 36,000 characters, the harness withholds it and asks the agent to print less or to use \texttt{exec\_qa}; without \texttt{exec\_qa}, it only asks the agent to print less. As Table~\ref{tab:no_exec_qa} shows, removing \texttt{exec\_qa} barely changes the average score and cost. Thus, the harness can be simplified further. 

\begin{table}[!h]
\centering
\resizebox{\linewidth}{!}{
\begin{tabular}{lccccc}
\toprule
 & EgoLifeQA & Week & Day & Month & Ego-R1-Bench \\
\midrule
\multicolumn{6}{l}{\textit{Score (\%)}} \\
With \texttt{exec\_qa} & 76.84 & 57.90 & 65.55 & 49.17 & 79.20 \\
Without \texttt{exec\_qa} & 78.20 & 61.25 & 64.75 & 48.15 & 74.00 \\
\midrule
\multicolumn{6}{l}{\textit{Cost/q (US\textcent)}} \\
With \texttt{exec\_qa} & 1.38 & 1.50 & 1.68 & 1.45 & 1.21 \\
Without \texttt{exec\_qa} & 1.38 & 1.40 & 1.65 & 1.34 & 1.26 \\
\bottomrule
\end{tabular}}
\caption{Scores and cost per question with and without \texttt{exec\_qa}, with DeepSeek (xhigh reasoning) on the full question sets. \textit{With \texttt{exec\_qa}} is the mean over the five runs in Table~\ref{tab:std}.}
\label{tab:no_exec_qa}
\end{table}

\section{Results Across Multiple Runs}
\label{sec:multi_runs}
To test how stable the results of our main experiment are, we run it four more times with identical settings and report the mean and standard deviation for each dataset, including the main run (Table~\ref{tab:std}). The standard deviation is at most 2.71 points, and on EgoLifeQA, MM-Lifelong Week, and Ego-R1-Bench, even the lowest of the five runs is above our MAGIC-Video rerun with DeepSeek in Table~\ref{tab:main}. 

\begin{table}[!h]
\centering
\small
\setlength{\tabcolsep}{4.5pt}
\begin{tabular}{lcccc}
\toprule
Dataset & First & Mean & Std & Range \\
\midrule
EgoLifeQA & 77.40 & 76.84 & 1.43 & 75.00--78.80 \\
Week      & 59.75 & 57.90 & 1.39 & 55.75--59.75 \\
Day       & 64.75 & 65.55 & 2.27 & 62.50--69.50 \\
Month     & 48.39 & 49.17 & 0.75 & 48.31--50.24 \\
Ego-R1-Bench & 78.00 & 79.20 & 2.71 & 76.00--84.00 \\
\bottomrule
\end{tabular}
\caption{Run-to-run variance of VideoJSON with DeepSeek (xhigh reasoning). Each benchmark is run five times with identical settings. \textit{First} is the run reported in Table~\ref{tab:main}, \textit{Mean} and \textit{Std} are the mean and population standard deviation over the five runs, and \textit{Range} is the minimum and maximum over the five runs.}
\label{tab:std}
\end{table}



\section{Bilingual Results}
\label{sec:bilingual}
In the main paper, we run our method using only Chinese. To provide a contextualized comparison against MAGIC-Video, we also run our method using bilingual questions and options. The results are shown in Table~\ref{tab:bilingual}.

\begin{table}[t]
\centering
\resizebox{\linewidth}{!}{
\begin{tabular}{lccccccc}
\toprule
 & Run 1 & Run 2 & Run 3 & Run 4 & Run 5 & Mean & Std \\
\midrule
Bilingual & 75.40 & 76.80 & 77.00 & 78.80 & 77.80 & 77.16 & 1.13 \\
Chinese   & 77.40 & 75.00 & 78.80 & 75.40 & 77.60 & 76.84 & 1.43 \\
\bottomrule
\end{tabular}
}
\caption{Bilingual Comparison of Our Method on EgoLifeQA.}
\label{tab:bilingual}
\end{table}

\section{System Prompt}
\label{sec:prompt}
\begin{tcolorbox}[enhanced, breakable, colback=blue!3!white, colframe=blue!45!black,
  boxrule=0.6pt, arc=2mm, left=2mm, right=2mm, top=1mm, bottom=1mm,
  fonttitle=\bfseries\small, title={System prompt of VideoJSON (EgoLifeQA)}]
\small
\textit{[The JSON Schema of the caption object of every index record.]}

\medskip
Answer the question from the video index at /data/minutes.json. The video is the first-person footage of \{name\}, who wears the camera, over several days; the questions are asked by \{name\}, so I, me, and my refer to \{name\}. The user message gives the day and the time at which the question is asked. The index is a JSON list of chronological one-minute clips. Each record has clip\_id, day (DAY1 to DAY7), start\_time and end\_time (wall-clock time of that day, HH:MM:SS), wall\_start\_s and wall\_end\_s (the same times in seconds since midnight), tick\_start\_s and tick\_end\_s (seconds of recorded footage within that day) and caption.

\medskip
The caption has dense\_caption (a detailed description of the minute in time order), events (what happens, one entry per occurrence, with start\_s, end\_s, type, who and description), unexpected (surprising or striking things with start\_s and description), entities (people, animals and objects with name, type, description and state), places (name and description), actions (who and a hierarchy from the overall activity down to the current step), relations (subject, relation, object), transcription (the cleaned transcript of the speech of the minute), explanation (the purpose of the minute in the whole video), on\_screen\_text (text and where) and mood. Seconds inside a caption are 0 to 59 of that minute.

\medskip
Use the exec tool to run ordinary Python or bash that loads the JSON and searches, filters and aggregates these fields. Be mindful about tool call output size and chunk the output when needed; the captions are very long. You have 30 model calls; after that you must answer with submit\_answer only.

\medskip
Each question comes with \{option\_count\} options. Give the final answer by calling submit\_answer with the letter of the option.
\end{tcolorbox}

\end{document}